\documentclass[10pt,journal]{IEEEtran}
\usepackage{graphicx}
\usepackage[english]{babel}
\usepackage{float}
\usepackage{amsthm, amsmath, amsfonts, mathtools, amssymb}
\usepackage{booktabs}
\usepackage{enumitem}
\usepackage{url}
\usepackage{geometry}
\usepackage[T1]{fontenc}
\usepackage[utf8]{inputenc}
\usepackage{lmodern}
\usepackage{algorithm}
\usepackage{algpseudocode}      
\usepackage[hidelinks]{hyperref}

\newcommand{\be}{\begin{equation}}
\newcommand{\ee}{\end{equation}}

\makeatletter
\algrenewcommand\algorithmiccomment[1]{\hfill\(\triangleright\)\,
  \parbox[t]{.55\linewidth}{\raggedright #1}}
\makeatother

\title{Comparing BFGS and OGR for Second-Order Optimization}
\author{
\IEEEauthorblockN{Adrian Przybysz\IEEEauthorrefmark{1}, Mikolaj Kolek\IEEEauthorrefmark{2}, Franciszek Sobota\IEEEauthorrefmark{3}, Jarek Duda\IEEEauthorrefmark{4}}
\IEEEauthorblockA{\IEEEauthorrefmark{1}Collegium Da Vinci in Poznan, Poland, Email: \texttt{adrian.j.przybysz@gmail.com}} 
\IEEEauthorblockA{\IEEEauthorrefmark{2}AGH University of Cracow, Poland} 
\IEEEauthorblockA{\IEEEauthorrefmark{3}University of Warsaw, Poland} 
\IEEEauthorblockA{\IEEEauthorrefmark{4}Faculty of Mathematics and Computer Science, Jagiellonian University, Crakow, Poland, \emph{dudajar@gmail.com}}}
\begin{document}
\maketitle

\begin{abstract}
Estimating the Hessian matrix, especially for neural network training, is a challenging problem due to high dimensionality and cost. In this work, we compare the classical Sherman–Morrison update used in the popular BFGS method (Broyden–Fletcher–Goldfarb–Shanno), which maintains a positive definite Hessian approximation under a convexity assumption, with a novel approach called Online Gradient Regression (OGR). OGR performs regression of gradients against positions using an exponential moving average to estimate second derivatives online, without requiring Hessian inversion. Unlike BFGS, OGR allows estimation of a general (not necessarily positive definite) Hessian and can thus handle non-convex structures. We evaluate both methods across standard test functions and demonstrate that OGR achieves faster convergence and improved loss, particularly in non-convex settings.

\end{abstract}
\textbf{Keywords:} machine learning, neural network training, optimization, Hessian, BFGS, convexity

\section{Introduction}
Second-order derivatives can improve the choice of the optimization step in Newton-like methods. However, applying them to neural networks is challenging: the Hessian has $O(D^2)$ entries for $D$ parameters, its reliable estimation is costly, and Newton-style updates still require (explicit or implicit) inversion with $O(D^3)$ worst-case complexity.

Although exact second derivatives can be obtained by automatic differentiation (higher-order backprop) in modern frameworks (e.g., PyTorch) \cite{pytorch}, the overhead becomes prohibitive at scale. Computing or materializing the full Hessian scales quadratically in memory and time, and using it in Newton steps entails solving linear systems or inverting $H$, which adds cubic cost. Even when using Hessian–vector products to avoid storing $H$, repeated products are still needed to approximate curvature, which increases wall-clock time. This motivates estimating curvature \emph{from the gradient stream alone}—data that we already compute at every iteration—so as to avoid explicit Hessian construction and inversion.

In this paper we compare (i) the classical curvature estimation from gradient differences used by quasi-Newton methods such as BFGS with (ii) an online approach, \emph{Online Gradient Regression} (OGR), which performs linear regression of gradients versus positions and maintains the estimator with an exponential moving average (EMA) that exponentially down-weights stale information \cite{Duda2019-Parabola,Duda2019-OGR}. Unlike BFGS, which enforces a positive-definite inverse Hessian approximation under a convexity assumption, OGR estimates unconstrained local curvature that can have negative eigenvalues. While this permits saddle-aware behavior in principle, our implementation uses eigenvalue regularization for numerical stability.

Concretely, OGR models the local relation $g(\theta)\!\approx\!H(\theta-p)$ and updates the required sufficient statistics online at a negligible cost relative to gradient computation. We consider both the 1D, momentum-guided variant based on online parabola fitting \cite{Duda2019-Parabola} and its multidimensional extension that tracks several statistically relevant directions with periodic (or online) re-orthogonalization to keep the Hessian estimate near-diagonal in the modeled subspace \cite{Duda2019-OGR}.

\textbf{Summary of results (brief).} Across standard test functions and ablations with/without line search, OGR variants match or outperform BFGS in final objective and step efficiency, with particular gains in nonconvex landscapes where saddle handling matters. Exact Hessians (via AD) are used only as an oracle baseline to evaluate estimation quality, not to form steps.

\section{Online Gradient Regression (OGR)}

Online Gradient Regression (OGR) is a second-order optimization framework that accelerates stochastic gradient descent (SGD) by \emph{online} least-squares regression of noisy gradients to infer local curvature and the distance to a stationary point \cite{Duda2019-OGR}. The central assumption is that, in a small neighborhood, the objective $F(\theta)$ is well-approximated by a quadratic model, so the gradient varies approximately linearly with the parameters. OGR maintains exponentially weighted statistics of recent $(\theta_t, g_t)$ pairs and updates a local model each iteration at negligible extra cost compared to computing the gradient itself \cite{Duda2019-Parabola,Duda2019-OGR}.

\subsection{Direct multivariate approach}\label{hesder}
In given time $T$, based on recent gradients $g^t\in{\mathbb{R}^d}$ and positions $\theta^t \in{\mathbb{R}^d}$ for $t<T$, we would like to locally approximate behavior with 2nd order polynomial using parametrization:
$$f(\theta)=h+\frac{1}{2}(\theta-p)^T H(\theta-p)\qquad\qquad \nabla f=H(\theta-p)$$
for Hessian $H\in\mathbb{R}^{d\times d}$ and $p\in \mathbb{R}^d$ position of saddle or extremum. 

For local behavior we will work on averages with weights $w^t$ further decreasing exponentially, defining averages:
\be \bar{v}\equiv \sum_{t<T} w^t v^t \quad\left(\propto \sum_{t<T} \beta^{T-t} v^t\right)\ee.
The mean square error to optimize becomes:
$$\arg\min_{H,p}\ \sum_{i,t} w^t \left(g^t_i-\sum_k H_{ik}(\theta^t_k-p_k)\right)^2 $$
getting necessary conditions. First for $\partial_{p_j}=0$:
$$\forall_{j}\quad \sum_{t,i} w^t \left(g^t_i-\sum_k H_{ik}(\theta^t_k-p_k)\right)H_{ij}=0 $$
$$ \forall_i\qquad \overline{g_i} - \sum_k H_{ik}\overline{\theta_k}+s\sum_k H_{ik}p_k=0 $$
\be \overline{g}=H\overline{\theta}-s\,Hp=H(\overline{\theta}-s\,p) \label{eq1}\ee
for $\overline{g}_i=\overline{g_i}$, $\overline{\theta}_i=\overline{\theta_i}$ averaged vectors, $s=\sum_t w^t$. 

For $\partial_{H_{ij}}=0$ we get:
$$\forall_{i,j}\quad \sum_t w^t (\theta^t_j-p_j)\left(g^t_i-\sum_k H_{ik}(\theta^t_k-p_k)\right)=0 $$
$$\overline{g_i\theta_j}-\overline{g_i}p_j=\sum_k H_{ik} \left(\overline{\theta_k\theta_j}-p_k\overline{\theta_j}-
\overline{\theta_k}p_j+s\,p_k p_j\right)$$
$$\overline{g\theta}-\overline{g}p^T=H\overline{\theta\theta}-Hp\overline{\theta}^T
-H(\overline{\theta}-sp)p^T$$
where the last is matrix equation with $\overline{g\theta}_{ij}=\overline{g_i\theta_j}$,  $\overline{\theta\theta}_{ij}=\overline{\theta_i\theta_j}$ averages. Substituting (\ref{eq1}) twice $(Hp=s^{-1}(H\overline{\theta}-\overline{g}))$ we get:
$$\overline{g\theta}\overset{(\ref{eq1})}{=} H\overline{\theta\theta}-Hp\overline{\theta}^T\overset{(\ref{eq1})}{=}
H\overline{\theta\theta}-s^{-1}(H\overline{\theta}-\overline{g})\overline{\theta}^T$$
$$s\overline{g\theta}=
sH\overline{\theta\theta}-H\overline{\theta}\overline{\theta}^T+\overline{g}\overline{\theta}^T$$
\be H=\left(s\overline{g\theta}-\overline{g}\,\overline{\theta}^T\right)
\left(s\overline{\theta\theta}-\overline{\theta}\,\overline{\theta}^T\right)^{-1}
\label{regm}\ee
Denominator is weighted covariance matrix, hence should be positive defined. Using
\be p=(\overline{\theta}- H^{-1}\overline{g})/s \label{regmp} \ee
we get the $\nabla f=0$ position: extremum or saddle of degree 2 polynomial modeling our function.\\

For online implementation, we update these statistics via exponential moving average with decay parameter $\beta\in(0,1)$:
$$\overline{\theta_i \theta_j}^{t+1}=\beta\,\overline{\theta_i \theta_j}^{t}+(1-\beta)\,\theta_i^t \theta_j^t$$
The mean parameter $s$ is tracked similarly: $s^{t+1} = \beta s^t + 1$, representing the effective sample size. At initialization, we set the covariance matrices $\overline{\theta\theta}$ and $\overline{g\theta}$ to the identity matrix for numerical stability.
To obtain the update direction, we compute $H^{-1}$ via eigendecomposition with eigenvalue regularization:
\be \theta^{t+1} = \theta^t - \alpha H^{-1}(\theta^t) g(\theta^t) \ee
where eigenvalues of $H$ are clipped by their absolute value to a minimum threshold $\epsilon$ (typically $10^{-12}$) before inversion. Additionally, we clip the total step size to a maximum L2 norm $\tau$ to prevent instability: if $\|\Delta\theta\| > \tau$, we rescale $\Delta\theta \leftarrow \tau \cdot \Delta\theta / \|\Delta\theta\|$.

\subsection{Symmetrized Hessian estimator}
The regression formula (\ref{regm}) does not generally produce a symmetric matrix. To ensure symmetry, we derive the solution assuming $H=H^T$ from the outset. Summing the necessity equations $\partial_{H_{ij}}+\partial_{H_{ji}}=0$ yields:
$$ \overline{\hat{g}\hat{\theta}^T} + \overline{\hat{\theta}\hat{g}^T}=
H\,\overline{\hat{\theta}\hat{\theta}^T}+\overline{\hat{\theta}\hat{\theta}^T}\,H $$
Let $\overline{\hat{\theta}\hat{\theta}^T}=ODO^T$ where $D=\textrm{diag}(\sigma^2_1,\ldots,\sigma^2_d)$ is the eigenvalue decomposition. Denoting $C=O^T\left(\overline{\hat{g}\hat{\theta}^T} + \overline{\hat{g}\hat{\theta}^T}^T\right) O$ and $H'=O^T H O$, the equation becomes:
$$C=DH'+H'D\qquad\Rightarrow\qquad C_{ij} = H'_{ij} (\sigma_i^2+\sigma_j^2)$$
Solving element-wise:
\be H=O\left[\left(O^T
\left(\overline{\hat{g}\hat{\theta}^T} + \overline{\hat{g}\hat{\theta}^T}^T\right) O\right)_{ij}/(\sigma_i^2+\sigma_j^2)  \right]_{ij}
O^T\label{hsf}\ee
with coordinate-wise division by $(\sigma_i^2+\sigma_j^2)$. This is the symmetrized estimator used in our experiments.

\subsection{Local quadratic and gradient regression}
\emph{Note: Sections 2.3--2.5 describe alternative 1D and low-dimensional formulations from the literature \cite{Duda2019-Parabola,Duda2019-OGR}. Our experiments (Section 4) use the full multivariate approach from Sections 2.1--2.2.}

In one dimension (a chosen coordinate or direction), a local quadratic
\[
F(\theta) \;\approx\; C \;+\; \tfrac{1}{2}\,\lambda\,(\theta - p)^2
\]
implies a linear gradient model
\[
g(\theta) \;\equiv\; \frac{dF}{d\theta} \;\approx\; \lambda\,(\theta - p).
\]
Given a stream $\{(\theta_t,g_t)\}$, OGR fits this line by weighted least squares (WLS). With exponentially decayed weights, the online (EMA) sums
\[
S_{\theta^2}=\sum_t w_t\,\theta_t^2,\;\;
S_{\theta g}=\sum_t w_t\,\theta_t g_t,\;\;
S_{\theta}=\sum_t w_t\,\theta_t,\;\;
S_{g}=\sum_t w_t\,g_t,\;\;
N=\sum_t w_t
\]
yield the closed-form WLS estimates \cite{Duda2019-Parabola,Duda2019-OGR}
\[
\hat{\lambda}\;=\;\frac{N\,S_{\theta g}-S_{\theta}\,S_{g}}{N\,S_{\theta^2}-S_{\theta}^2},
\qquad
\hat{p}\;=\;\frac{S_{\theta}\,\hat{\lambda}-S_{g}}{\hat{\lambda}\,N}
\;=\;\bar{\theta}\;-\;\frac{\bar{g}}{\hat{\lambda}},
\]
where $\bar{\theta}=S_{\theta}/N$ and $\bar{g}=S_{g}/N$. Numerically, one clips $|\hat{\lambda}|$ away from zero (or caps $|\hat{p}-\theta|$) to avoid unstable jumps on flat/inflection regions \cite{Duda2019-Parabola}.

\subsection{Online update rule (1D)}
Once $(\hat{\lambda},\hat{p})$ are available, OGR takes a trust-weighted Newton-like step toward the predicted zero-gradient point:
\[
\theta_{t+1}\;=\;\theta_t \;+\; \alpha\,\mathrm{sign}(\hat{\lambda})\,\big[\hat{p}-\theta_t\big]_{\text{clipped}},
\]
with $0<\alpha\le1$ controlling confidence in the local model. For $\hat{\lambda}>0$ (convex), this moves toward the estimated minimum; for $\hat{\lambda}<0$ (concave/saddle along the chosen direction), the sign factor inverts the step to repel from the unstable extremum \cite{Duda2019-OGR}. A short warm-up (a few SGD steps) can initialize the EMA statistics before trusting the regression.

\subsection{Multidimensional OGR}
Directly estimating a full $D\times D$ Hessian is infeasible. OGR therefore restricts second-order modeling to a low-dimensional subspace spanned by a small set of \emph{statistically relevant directions} extracted from recent gradients \cite{Duda2019-OGR}. Let $V=[v_1,\dots,v_d]\in\mathbb{R}^{D\times d}$ be an orthonormal basis ($d\ll D$). Project parameters and gradients to coordinates $\theta_{i}=v_i^\top\theta$, $g_{i}=v_i^\top g$. Maintain the same EMA statistics per direction $i$ and compute
\[
\hat{\lambda}_i \;=\; \frac{N\,S_{\theta_i g_i}-S_{\theta_i}\,S_{g_i}}{N\,S_{\theta_i^2}-S_{\theta_i}^2},
\qquad
\hat{p}_i \;=\; \bar{\theta}_i - \frac{\bar{g}_i}{\hat{\lambda}_i}.
\]
The multidimensional update then aggregates independent 1D steps in the subspace:
\[
\theta_{t+1}\;=\;\theta_t \;+\; \alpha \sum_{i=1}^d \mathrm{sign}(\hat{\lambda}_i)\,\big(\hat{p}_i-\theta_{i}\big)\,v_i,
\]
optionally adding a conservative first-order step in the orthogonal complement. Periodically, one re-orthogonalizes/rotates $V$ so the modeled Hessian remains nearly diagonal in the subspace \cite{Duda2019-OGR}.

\section{BFGS}
\subsection{Newton and Quasi-Newton Methods}

The Newton method is a second-order optimization algorithm that uses both the gradient and the Hessian matrix to find a minimum of the objective function. It locally approximates the function via second-order Taylor expansion:
\[
f(\mathbf{x}) \approx f(\mathbf{x}_k) + \nabla f(\mathbf{x}_k)^T(\mathbf{x} - \mathbf{x}_k) + \frac{1}{2}(\mathbf{x} - \mathbf{x}_k)^T \nabla^2 f(\mathbf{x}_k)(\mathbf{x} - \mathbf{x}_k)
\]

The update rule becomes:
\[
\mathbf{x}_{k+1} = \mathbf{x}_k - [\nabla^2 f(\mathbf{x}_k)]^{-1} \nabla f(\mathbf{x}_k)
\]

While Newton's method often converges rapidly, it is limited by the computational cost of computing and inverting the Hessian matrix. These operations scale as $\mathcal{O}(d^2)$ in memory and $\mathcal{O}(d^3)$ in time, where $d$ is the number of parameters. This makes the method impractical for high-dimensional problems such as deep learning.

To overcome these limitations, \textbf{quasi-Newton methods} approximate the Hessian matrix (or its inverse) iteratively, without computing it explicitly. The update rule becomes:
\[
\mathbf{x}_{k+1} = \mathbf{x}_k - B_k^{-1} \nabla f(\mathbf{x}_k)
\]
where $B_k$ is the approximation of the Hessian (or its inverse) at iteration $k$.

Quasi-Newton methods strike a balance between the fast convergence of Newton's method and the lower computational burden of first-order methods. Among them, the BFGS algorithm is one of the most popular and effective variants, which we describe next.

\subsection{BFGS algorithm}
The BFGS algorithm uses a quasi-Newton equation to ap-
proximate the Hessian matrix or its inverse and uses symmetric
second-order updates to maintain its positive definiteness, thanks
to which the algorithm is incredibly quick to converge. The matrix update ensures that it remains positive definite
and attempts to accurately detect the curvature of the function
we are approximating. BFGS doesn't need to explicitly compute the Hessian matrix because of using iterative updates of its initial approxima-
tion using corrections that are computed as differences between
successive steps and their gradients. The formula updating
the approximation of the inverse Hessian matrix is crucial
in this implementation because it allows for computationally
efficient updating without the need to invert the matrix. This is
a key moment in the implementation because matrix inversion
is very computationally inefficient. 
The formula is as follows:
\[
{\displaystyle B_{k+1}^{-1}=\left(I-{\frac {\mathbf {s} _{k}\mathbf {y} _{k}^{T}}{\mathbf {y} _{k}^{T}\mathbf {s} _{k}}}\right)B_{k}^{-1}\left(I-{\frac {\mathbf {y} _{k}\mathbf {s} _{k}^{T}}{\mathbf {y} _{k}^{T}\mathbf {s} _{k}}}\right)+{\frac {\mathbf {s} _{k}\mathbf {s} _{k}^{T}}{\mathbf {y} _{k}^{T}\mathbf {s} _{k}}}.}
\]
where:
\begin{itemize}
    \item $B_{k}^{-1}$ is the approximation of the inverse Hessian matrix at iteration $k$.
    \item $\mathbf{s}_k = \mathbf{x}_{k+1} - \mathbf{x}_k$ is the change in the position.
    \item $\mathbf{y}_k = \nabla f(\mathbf{x}_{k+1}) - \nabla f(\mathbf{x}_k)$ is the change in the gradient.
    \item $I$ is the identity matrix.
\end{itemize}

\subsection{Limitations and extensions}
The BFGS algorithm incurs high computational costs in each iteration because it stores the matrix or its inverse in order to update it in subsequent steps. Comparisons have shown that the amount of memory needed to perform calculations using the BFGS algorithm is O(d²) \cite{Jin2022}, which is significantly more than the gradient descent method, which requires O(d) memory. The BFGS algorithm compensates for the high computational costs with accelerated convergence in strongly convex, smooth conditions. 

One possible solution to this problem is to use Limited-memory BFGS \cite{Nocedal1980}, which does not store the full matrix, allowing this method to be used in more complex problems.

BFGS often requires fewer iterations than the gradient descent method, as confirmed by experiments conducted in Block BFGS Methods \cite{Gao2016}.

Research shows that when the BFGS algorithm is applied to non-convex problems in machine learning, modifications are necessary due to computational and numerical issues. For example, variants that use limited memory help manage high memory requirements. The variance reduction, damping, and regularization techniques introduced in the study \cite{Liu2022}) mitigate numerical instability by smoothing noisy gradient estimates and improving curvature approximation. In summary, the basic BFGS algorithm has limitations in terms of both computational efficiency and numerical robustness when applied to non-convex machine learning tasks.

Below are several situations in which the BFGS algorithm fails to correctly approximate the Hessian matrix or its inverse. 
In experiments with standard line search, disturbed gradients
and inaccurate search conditions can cause irregular step sizes,
which in turn can lead to distortion of the curvature estimate and negatively
affect the convergence of the algorithm [5].
We need to be careful with two more things. One, tiny steps and rounding errors can make our Hessian matrix approximation completely unreliable, even for simple functions \cite{Zhang2020}. Two, how we set up the algorithm matters a lot. A good starting point and smart update rules will make it converge much faster and more reliably \cite{Jin2024}.

\section{Comparative Analysis of BFGS and OGR Convergence under Varying Conditions}

\subsection{Test Functions and Methodology}

We have conducted our experiments on the set of many single objective functions taken mostly from \cite{oldenhuis2012testfunctions}. 
We chose them for their diversity, some are convex, some have narrow curved valleys, and others have many local minima. This set of test functions allows us to thoroughlt test our optimizer. 
Below are formulas of the functions we used:

\begin{itemize}[leftmargin=*,label=\bfseries{\textbullet},font=\small]
    \item \textbf{Sphere function:} $f(\mathbf{x}) = \sum_{i=1}^d x_i^2$
    \item \textbf{Rosenbrock function:} $f(\mathbf{x}) = \sum_{i=1}^{d-1} [100(x_{i+1} - x_i^2)^2 + (1 - x_i)^2]$
    \item \textbf{Rastrigin function:} $f(\mathbf{x}) = 10d + \sum_{i=1}^d [x_i^2 - 10 \cos(2\pi x_i)]$
    \item \textbf{Ackley function:}
    \begin{equation*}
        \begin{split}
            f(\mathbf{x}) = &-20 \exp\left(-0.2\sqrt{\frac{1}{d}\sum_{i=1}^d x_i^2}\right) \\
            &- \exp\left(\frac{1}{d}\sum_{i=1}^d \cos(2\pi x_i)\right) + 20 + e
        \end{split}
    \end{equation*}
    \item \textbf{Griewank function:} $f(\mathbf{x}) = \sum_{i=1}^d \frac{x_i^2}{4000} - \prod_{i=1}^d \cos\left(\frac{x_i}{\sqrt{i}}\right) + 1$
    \item \textbf{Schwefel function:} $f(\mathbf{x}) = 418.9829d - \sum_{i=1}^d x_i \sin(\sqrt{|x_i|})$
    \item \textbf{Zakharov function:} $f(\mathbf{x}) = \sum_{i=1}^d x_i^2 + \left(\sum_{i=1}^d 0.5ix_i\right)^2 + \left(\sum_{i=1}^d 0.5ix_i\right)^4$
    \item \textbf{Himmelblau's function:} $f(x, y) = (x^2 + y - 11)^2 + (x + y^2 - 7)^2$
    \item \textbf{Beale function:} $f(x, y) = (1.5 - x + xy)^2 + (2.25 - x + xy^2)^2 + (2.625 - x + xy^3)^2$
\end{itemize}

We performed tests with 64-bit precision to ensure greater optimization accuracy, limiting each test to 2000 steps. For each function, we used 200 random points taken from uniform distributions within the limits specified for each function. We sorted the results by final loss in ascending order.

\subsection{Optimizer Configuration}

We tested the OGR and BFGS optimizers in two configurations: with line search and without line search. We set the step length parameter for both optimizers to the same value depending on the function. For the Rastrigin function it was 0.6 and for the rest of the functions it was 0.5. This parameter is used by optimizers that do not use line search, while optimizers that use line search use it to select the appropriate step length. 
For OGR, we use the full multivariate formulation with the symmetrized Hessian estimator from equation (\ref{hsf}). The EMA decay parameter $\beta$ was set to 0.2, a value selected experimentally. For numerical stability, we initialize the parameter and gradient covariance matrices to the identity matrix at the start of optimization. The eigenvalue clipping parameter was set to $\epsilon = 10^{-12}$. The Hessian inverse is computed via eigendecomposition with eigenvalues clipped by their absolute value to this threshold before inversion. This prevents numerical instability near flat regions while preserving information about negative curvature. The OGR optimizer also uses a parameter that limits the maximum step length (L2 norm) to reduce learning instability. A detailed description of the implementation and parameters of line search can be found in the section below.

The starting points for our optimizers were taken from uniform distributions within bounds specified separately for each function. Bounds for individual functions:

\begin{itemize}
    \item \textbf{Sphere:} $x_i \in [-5.0, 5.0]$
    \item \textbf{Rosenbrock:} $x_i \in [-2.0, 2.0]$
    \item \textbf{Rastrigin:} $x_i \in [-5.12, 5.12]$
    \item \textbf{Ackley:} $x_i \in [-5.0, 5.0]$
    \item \textbf{Griewank:} $x_i \in [-5.0, 5.0]$
    \item \textbf{Schwefel:} $x_i \in [-500.0, 500.0]$
    \item \textbf{Zakharov:} $x_i \in [-5.0, 5.0]$
    \item \textbf{Himmelblau:} $x \in [-5.0, 5.0]$, $y \in [-5.0, 5.0]$
    \item \textbf{Beale:} $x \in [-4.5, 4.5]$, $y \in [-4.5, 4.5]$
\end{itemize}

\begin{figure}[H]
    \centering
    \includegraphics[width=0.4\textwidth]{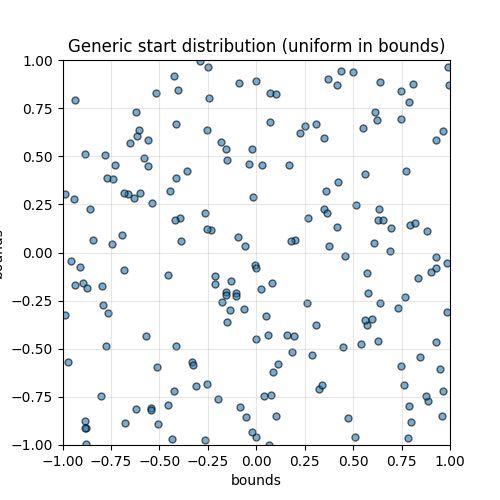}
    \caption{Distribution of starting points for a 2D function. All points were randomly sampled from a uniform distribution within the specified bounds, ensuring consistency across all tests.}
    \label{fig:starting_points}
\end{figure}

The starting points for all functions have a consistent distribution, as illustrated in Figure \ref{fig:starting_points}.

\newpage
\begin{figure*}[t]
    \centering
    \includegraphics[width=\textwidth]{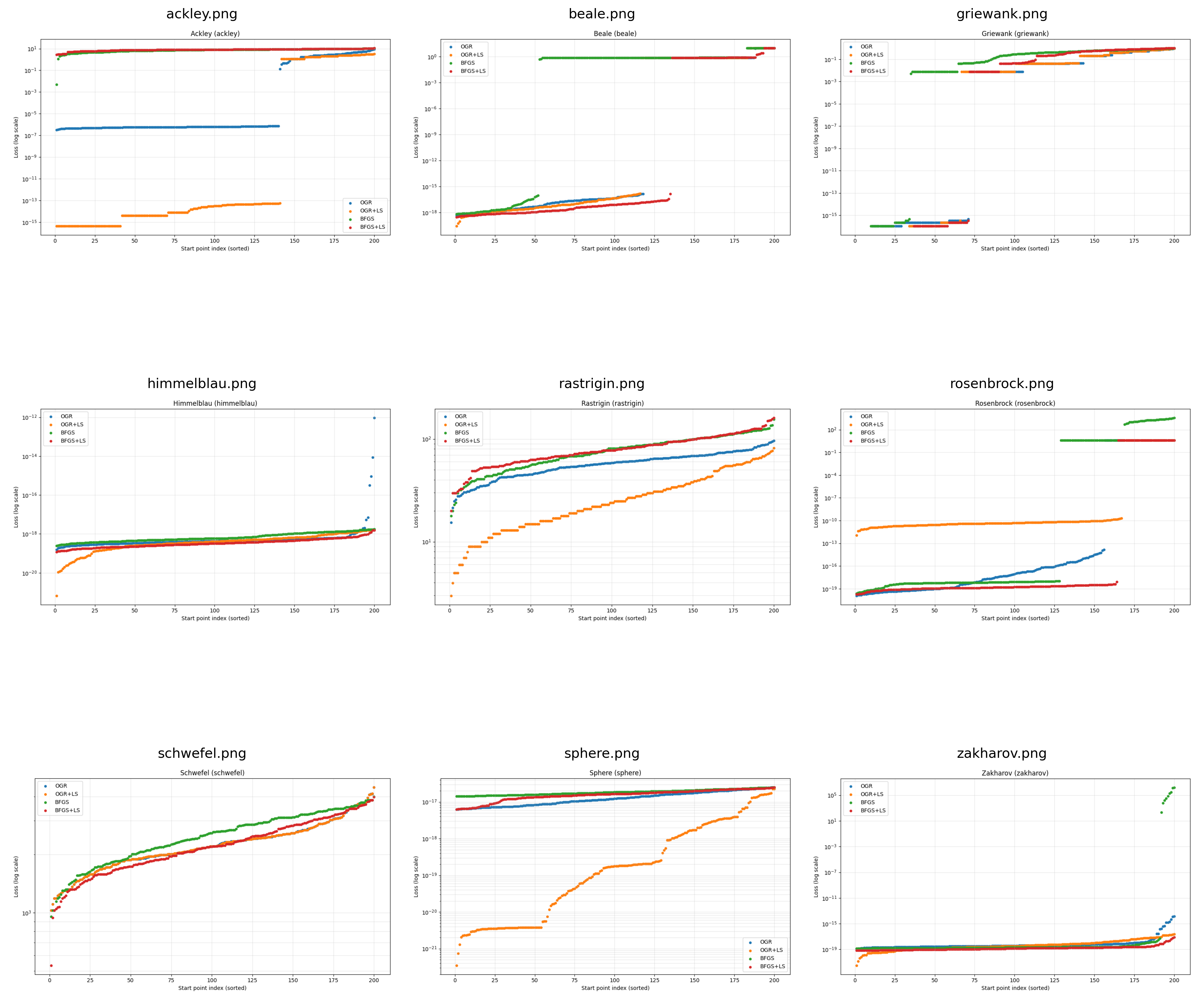}
    \caption{Performance comparison of BFGS and OGR with and without a line search procedure. The plots show the final loss distribution for 200 random starting points on various test functions.}
    \label{fig:results_plots}
\end{figure*}

\subsection{Results and Discussion}

The results of our experiments are presented in Figure \ref{fig:results_plots}.

Based on the tests and the above graphs, we draw the following conclusions:

\begin{itemize}
\item \textbf{Performance with Line Search:} The OGR optimizer with line search wins in most benchmarks. In all tests, OGR with the line search performs better or as well as the BFGS optimizer with the line search. 

\item \textbf{Performance without Line Search:} The OGR optimizer without line search outperforms the BFGS optimizer without line search in all tests.
\end{itemize}

In the case of the Rosenbrock function, our experiments showed that OGR without line search performs better than OGR with line search. We suspect that this counterintuitive result stems from the presence of a narrow curved valley in the Rosenbrock function, causing line search to select overly slow steps, while OGR without line search moves with a fixes step size toward a global minimum. 

Our experiments show that OGR performs significantly better with and without line search.

\newpage
\subsection{Line Search implementation}

 The line search we used in our experiments is based on a backward line search algorithm that uses the Armijo \cite{Armijo1966} condition to find the appropriate step size. Armijo condition: $$f(\mathbf{x}_k + \alpha_k \mathbf{p}_k) \leq f(\mathbf{x}_k) + c \alpha_k \nabla f(\mathbf{x}_k)^T \mathbf{p}_k$$ Where: \begin{itemize} \item $\alpha_k$: Step size \item $\mathbf{p}_k$: Search direction \item $c$: Constant that controls the required decrease in the function value \item $\nabla f(\mathbf{x}_k)^T \mathbf{p}_k$: The directional derivative of the function \end {itemize} 
\vspace{2em}

\begin{algorithm}[!t]
\caption{Backtracking line search with Armijo decrease}
\footnotesize
\begin{algorithmic}[1]
    \Require objective $f$, point $x$, direction $d$, optional gradient $g$
    \Require constants $c\in(0,1)$, $\tau\in(0,1)$, max backtracks $K_{\max}$
    \Ensure step $\Delta x$
    \State $f_0 \gets f(x)$
    \Statex \(\triangleright\) \textbf{Slope along $d$:} use $g$ if available; otherwise finite difference.
    \If{$g$ provided} \State $slope \gets g^\top d$
    \Else
      \State $\varepsilon \gets 10^{-8}$;\quad $n \gets \|d\|$
      \If{$n=0$} \State \Return $\mathbf{0}$ \Comment{zero direction}
      \EndIf
      \State $\delta \gets \varepsilon/\max(n,1)$
      \State $slope \gets \big(f(x+\delta d)-f_0\big)/\delta$
    \EndIf
    \Statex \(\triangleright\) \textbf{Guard:} if no descent, fall back to a simple decrease test.
    \State $useArmijo \gets (slope < 0)$
    \State $\alpha \gets 1$;\quad $best\_f \gets \text{None}$;\quad $best\_\Delta \gets \text{None}$
    \For{$t=1$ to $K_{\max}$}
      \State $f_{new} \gets f(x+\alpha d)$
      \If{$useArmijo$ \textbf{and} $f_{new} \le f_0 + c\,\alpha\,slope$} 
         \State \Return $\alpha d$ \Comment{Armijo satisfied}
      \ElsIf{\textbf{not} $useArmijo$ \textbf{and} $f_{new} < f_0$}
         \State \Return $\alpha d$ \Comment{simple decrease}
      \EndIf
      \If{$best\_f=\text{None}$ \textbf{or} $f_{new}<best\_f$}
         \State $best\_f \gets f_{new}$;\quad $best\_\Delta \gets \alpha d$
      \EndIf
      \State $\alpha \gets \tau \alpha$ \Comment{backtrack}
    \EndFor
    \State \Return $best\_\Delta$ \textbf{if} $best\_\Delta\neq\text{None}$ \textbf{else} $10^{-12} d$
\end{algorithmic}
\end{algorithm}

Below we describe how the algorithm works:
\begin{itemize}
  \item \textbf{Initialization:} The algorithm starts with an initial step size $\alpha_k = 1$.
  \item \textbf{Backtracking loop:} At each iteration, it evaluates the objective at $x_k + \alpha_k p_k$
        and checks whether the Armijo condition is satisfied.
  \item \textbf{Step size reduction:} If the condition is not satisfied, the step size $\alpha_k$ is reduced
        by a factor $\tau$ (set to $0.5$ in our experiments).
  \item \textbf{Termination:} The loop continues until a step size $\alpha_k$ satisfying the condition is found
        or the maximum number of backtracking steps (50 in our implementation) is reached.
\end{itemize}

\newpage
\section{Comparison of Optimization Trajectories of OGR and BFGS}

In this section we present the results of a comparison of the convergence of the OGR and BFGS optimizers on a set of test functions (Beale, Griewank, Zakharov, Sphere). 

\subsection{Experiment configuration}
\begin{itemize}
\item \textbf{Dimensionality:} All experiments were conducted in two dimensions ($d=2$).
\item \textbf{Restarts:} Each optimization process was repeated with 10 random initializations, and then the one that gave the best result was selected.
\item \textbf{Steps:} We set the maximum number of steps to 2000.
\item \textbf{Precision:} Calculations were performed with \texttt{float64} precision.
\item \textbf{Line search:} All experiments were performed without line search.
\end {itemize}

\begin{figure}[H]
\centering
\includegraphics[width=\textwidth]{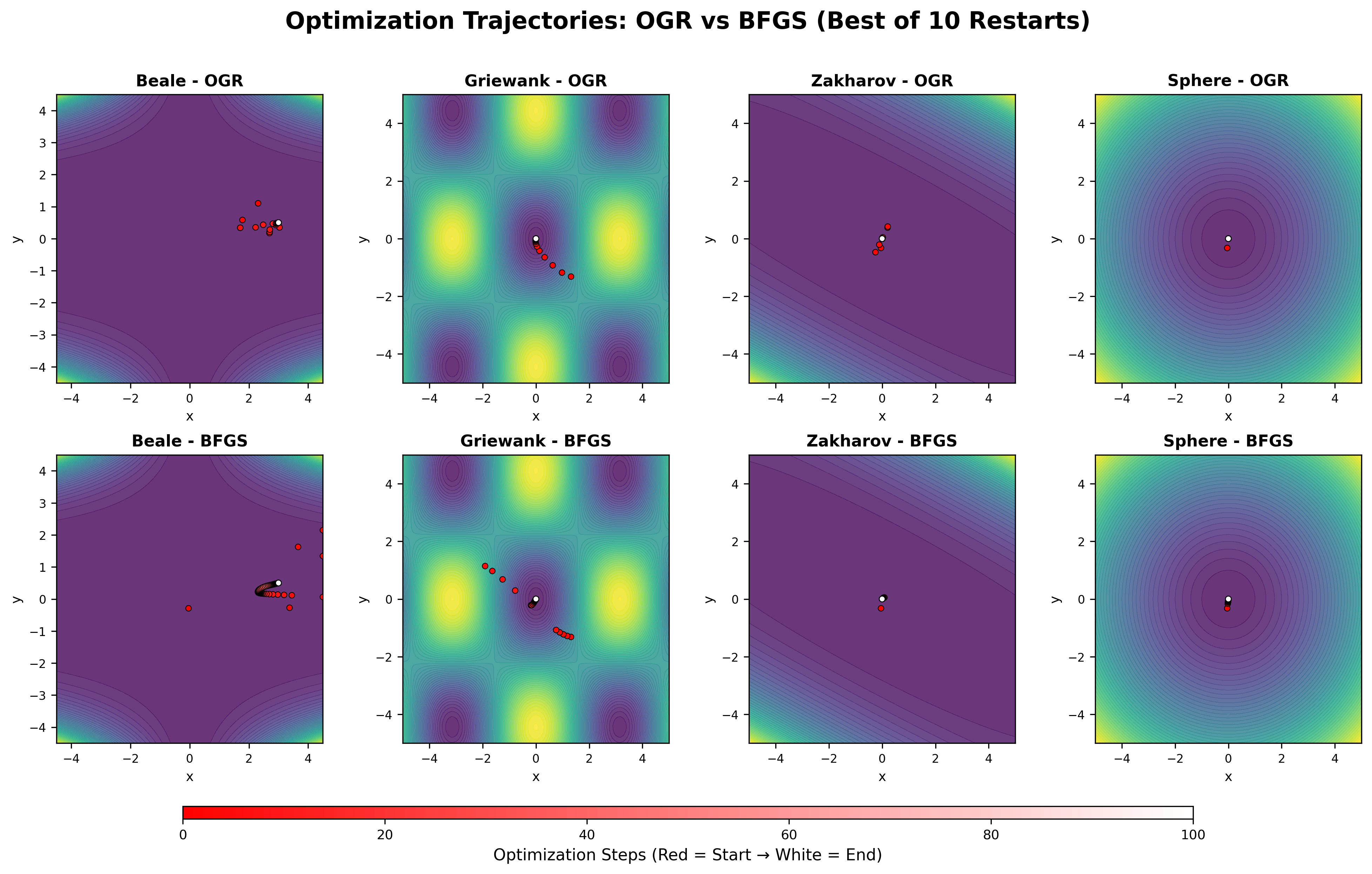}
\caption{Comparison of optimization for BFGS (top) and OGR (bottom) 
on selected test functions. OGR converges faster to the global minimum in all cases.}
\label{fig:ogr-vs-bfgs}
\end{figure}

Figure~\ref{fig:ogr-vs-bfgs} shows the results of our experiment. The top row shows the results for the BFGS optimizer, and the bottom row shows the results for the OGR optimizer. Each column shows a different test function. 
\clearpage

\clearpage

\subsection{Summary}
Our experiments show that OGR converges to the global minimum much faster than the BFGS optimizer. In the test functions both optimizers achieve a result very close to the global minimum, but OGR does so in significantly fewer steps. We can therefore assume that the OGR optimizer makes more effective use of curvature information about the curvature of the function, which allows it to converge to the minimum of the function more quickly.



\section{Implementation Environment}
The implementation of both algorithms was carried out in Python with use of Pytorch \cite{pytorch} framework, making it suitable for future experiments in neural network research. To ensure the numerical stability of second-order methods we used \texttt{float64} precision in all calculations. To ensure the repeatability during experiments we used a fixed random seed value which was 42.





\section{Conclusion}
OGR provides second-order curvature guidance online by regressing recent gradients against positions, thereby avoiding explicit Hessian computation or inversion. This adaptive local model guides iterates toward minima in directions of positive curvature and repels them from saddles along negative curvature, which is crucial for navigating non-convex landscapes. In contrast, classical BFGS maintains a positive-definite inverse Hessian approximation (requiring $O(D^2)$ memory), making it effective on smooth, strongly convex problems but increasingly restrictive in very high dimensions or around non-convex structures.

Across standard benchmarks, OGR—both with and without line search—typically achieved lower final losses in fewer steps than BFGS. A notable exception is the Rosenbrock function: its narrow curved valley caused the line search to become overly conservative, so that the fixed-step OGR variant actually outperformed the line-searched version. These experiments suggest that OGR makes more effective use of curvature information and can converge faster to minima, especially in challenging optimization landscapes.

Practically, OGR delivers much of the benefit of second-order methods at near first-order cost, making it a strong default when exact Hessians are impractical or too costly to compute. Its limitations include sensitivity to hyperparameters such as the EMA decay rate, step clipping near flat curvature, and the choice or refresh rate of the modeled subspace. Periodic subspace re-orthogonalization helps mitigate these issues but do not fully eliminate them. 

Future work will apply OGR at scale to neural network training, combine it with momentum/Adam-style adaptivity, and explore automated schedules for EMA decay and subspace selection.

\bibliographystyle{IEEEtran}
\bibliography{cites}

\end{document}